\documentclass[runningheads]{llncs}
\usepackage[T1]{fontenc}
\usepackage{graphicx,verbatim}
\usepackage{graphicx}
\usepackage{subcaption}  
\usepackage{hyperref}
\begin{document}
\title{Veriserum: A dual-plane fluoroscopic dataset with knee implant phantoms for deep learning in medical imaging}
\author{
Jinhao Wang \thanks{This work has been accepted at MICCAI 2025.}
 \and
Florian Vogl \and
Pascal Schütz \and 
Saša Ćuković \and \\
William R.~Taylor\thanks{Corresponding author: bt@ethz.ch}
}


\authorrunning{J. Wang et al.}

\institute{ETH Zürich, Laboratory for Movement Biomechanics, Institute for Biomechanics, Switzerland\\
\email{jinhao.wang@hest.ethz.ch}}


\maketitle              
\begin{abstract}
Veriserum is an open-source dataset designed to support the training of deep learning registration for dual-plane fluoroscopic analysis. It comprises approximately 110,000 X-ray images of 10 knee implant pair combinations (2 femur and 5 tibia implants) captured during 1,600 trials, incorporating poses associated with daily activities such as level gait and ramp descent. Each image is annotated with an automatically registered ground-truth pose, while 200 images include manually registered poses for benchmarking.

Key features of Veriserum include dual-plane images and calibration tools. The dataset aims to support the development of applications such as 2D/3D image registration, image segmentation, X-ray distortion correction, and 3D reconstruction. Freely accessible, Veriserum aims to advance computer vision and medical imaging research by providing a reproducible benchmark for algorithm development and evaluation. The Veriserum dataset used in this study is publicly available via \url{https://movement.ethz.ch/data-repository/veriserum.html}, with the data stored at ETH Zürich Research Collections: \url{https://doi.org/10.3929/ethz-b-000701146}.



\end{abstract}
\section{Introduction}
2D/3D image registration based on X-ray imaging is a key requirement for various biomechanical and medical applications, ranging from joint research to surgery guidance systems~\cite{Chen2024,Guan2016,Postolka2022}. Traditional registration algorithms typically rely on digitally reconstructed radiographs (DRRs), or rendering, which matches a rendered image of the 3D model to the target image by maximizing their similarity via optimization or manual vision~\cite{Flood2018,Petersen2023,Tsai2011,}. While this approach has been effective, it faces computational complexity and robustness challenges, especially in clinical settings. Recently, deep learning approaches have achieved considerable success in image registration tasks by leveraging large datasets~\cite{Li2018,Stevsic2021}. However, their application in X-ray image registration is constrained by the scarcity of high-quality medical data and stringent legal restrictions on data usage.

X-ray imaging presents practical challenges for registration purposes, including geometric distortion, limited resolution, and the precise measurement of source-intensifier distances ~\cite{Postolka2020}. Addressing these factors requires specialized calibration and distortion correction techniques, often lacking standardized datasets for benchmarking their performance. Despite these challenges, combining the strengths of deep learning and traditional registration methods plausibly holds strong potential to establish a robust, automatic 2D/3D registration pipeline~\cite{Gopalakrishnan2023,Jensen2023,Vogl2022,Zhang2023}. This pipeline requires large-scale, high-quality datasets that incorporate realistic poses and provide ground-truth data for model evaluation. 

Existing single-plane fluoroscopic datasets provide accurate pose labels for hip anatomy (DeepFluoro)  \cite{Grupp2020} and for knee implants in total knee arthroplasty (CAMS-Knee, Stan) \cite{Taylor2017,Dreyer2022} where precise kinematics are available.  However, dual-plane fluoroscopy has gained increasing attention over single-plane imaging due to its superior out-of-plane accuracy \cite{Guan2016}. Despite its advantages, an open-source dual-plane fluoroscopic dataset for knee joint image registration that not only provides accurate kinematics but also addresses data privacy issues remains an unmet need. While many studies evaluate their algorithms on existing datasets, most research pipelines still rely on proprietary, unpublished code and resources, limiting reproducibility and comparability in the field.

In this work, we introduce Veriserum, a unique dataset of 110,990 dual-plane X-ray images of tibial and femoral knee implants in various poses associated with activities of daily living. Using a precision robot, we simulated realistic movement patterns by positioning modified implants to predefined positions based on motion data from patients with knee implants. Alongside the X-ray images, the dataset includes raw calibration phantom measurements and calibration functions to enable distortion correction and the calculation of source-to-intensifier distances. 


By including dual-plane fluoroscopic images with tools for calibration correction, our dataset aims to a) provide the biomechanical research community with a high-quality dataset and ground-truth data to benchmark 2D/3D image registration techniques, b) support the computer vision community in developing novel automated methods for 2D/3D registration \cite{Wang2021}, image segmentation \cite{Vogl2022}, or 3D reconstruction \cite{mescheder2019occupancynetworkslearning3d} specific to knee implants, and c) facilitate the development of clinically relevant applications, such as kinematics assessment and surgical guidance systems, by enabling a practical and reproducible image registration pipeline. Through these contributions, Veriserum provides a robust dataset for advancing research at the intersection of biomechanics, computer vision, and medical imaging.

\section{Method}
\begin{figure}[htb]
    \centering
    \includegraphics[width=0.95\linewidth]{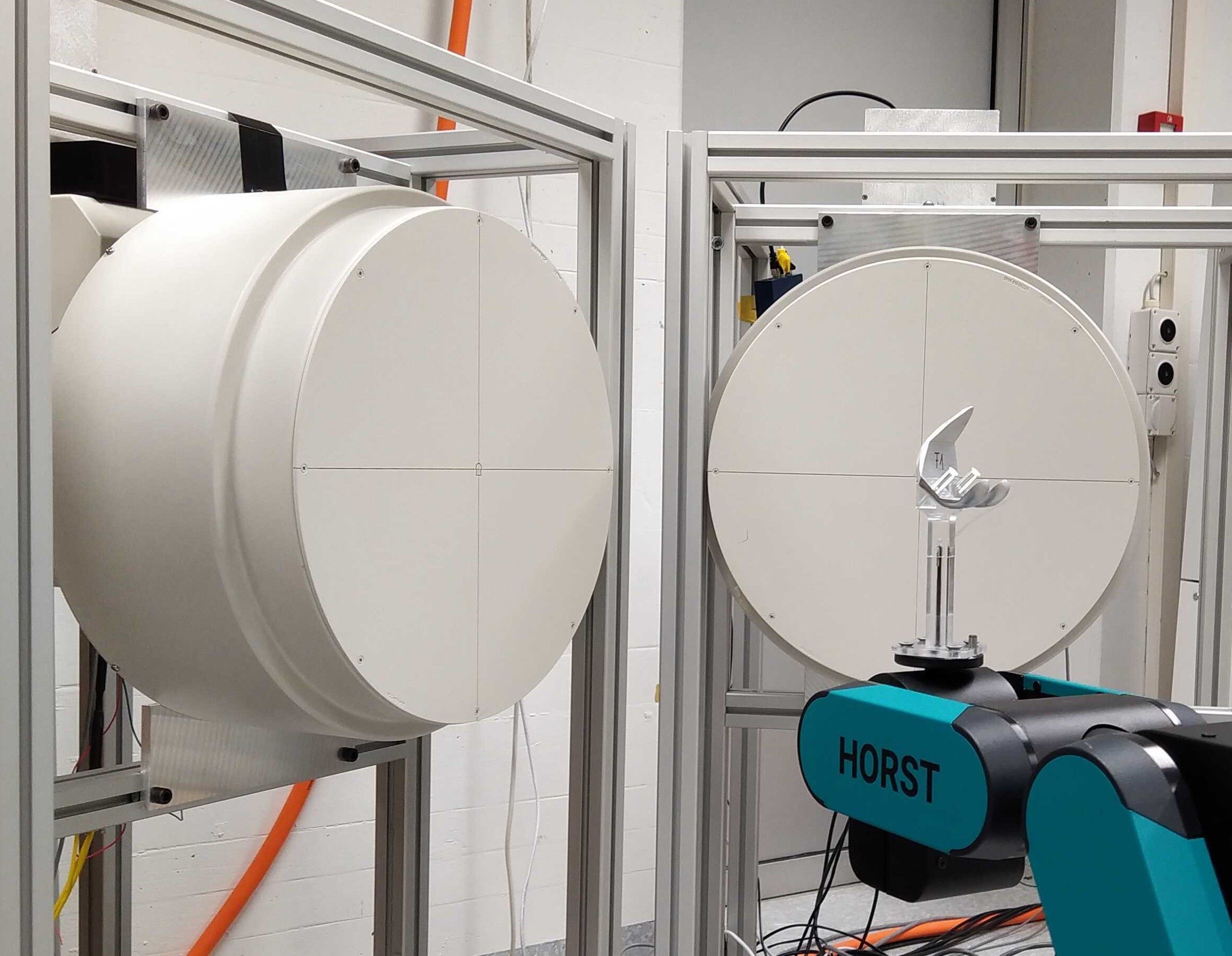}
    \caption{Schematic of the dual-plane X-ray imaging setup.}
    \label{fig:dual_plane}
\end{figure}

The experimental setup consists of a dual-plane fluoroscopic system and a precision robotic system with an implant model attached to the end-effector (Fig.\ref{fig:dual_plane}). This section describes the necessary elements involved in the data acquisition process.

\begin{figure}[htb]
    \centering
    \begin{subfigure}{0.48\linewidth}
        \centering
        \includegraphics[width=\linewidth]{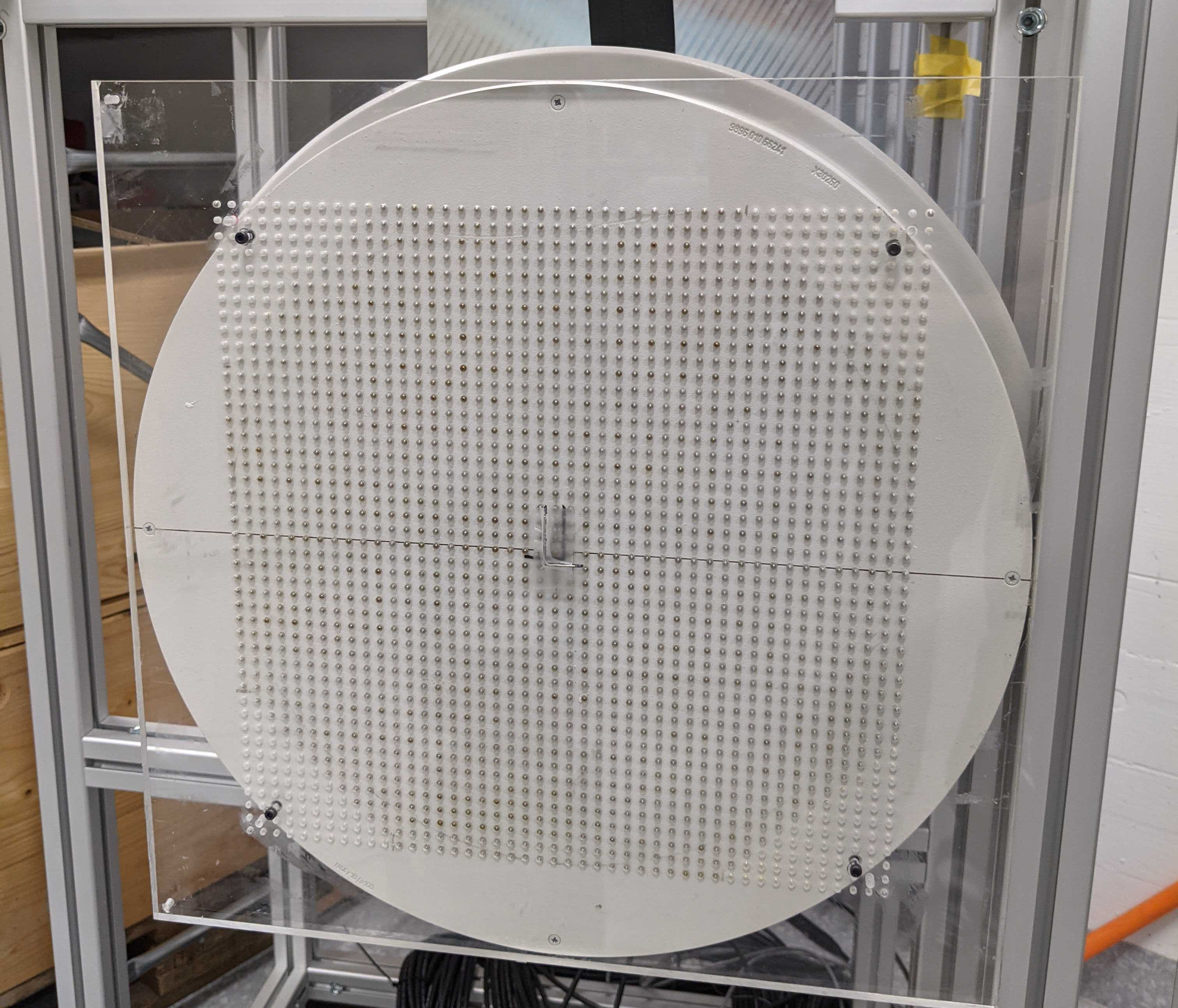}
        \caption{Bead grid calibration pattern.}
        \label{fig:bead_grid}
    \end{subfigure}
    \hfill
    \begin{subfigure}{0.48\linewidth}
        \centering
        \includegraphics[width=\linewidth]{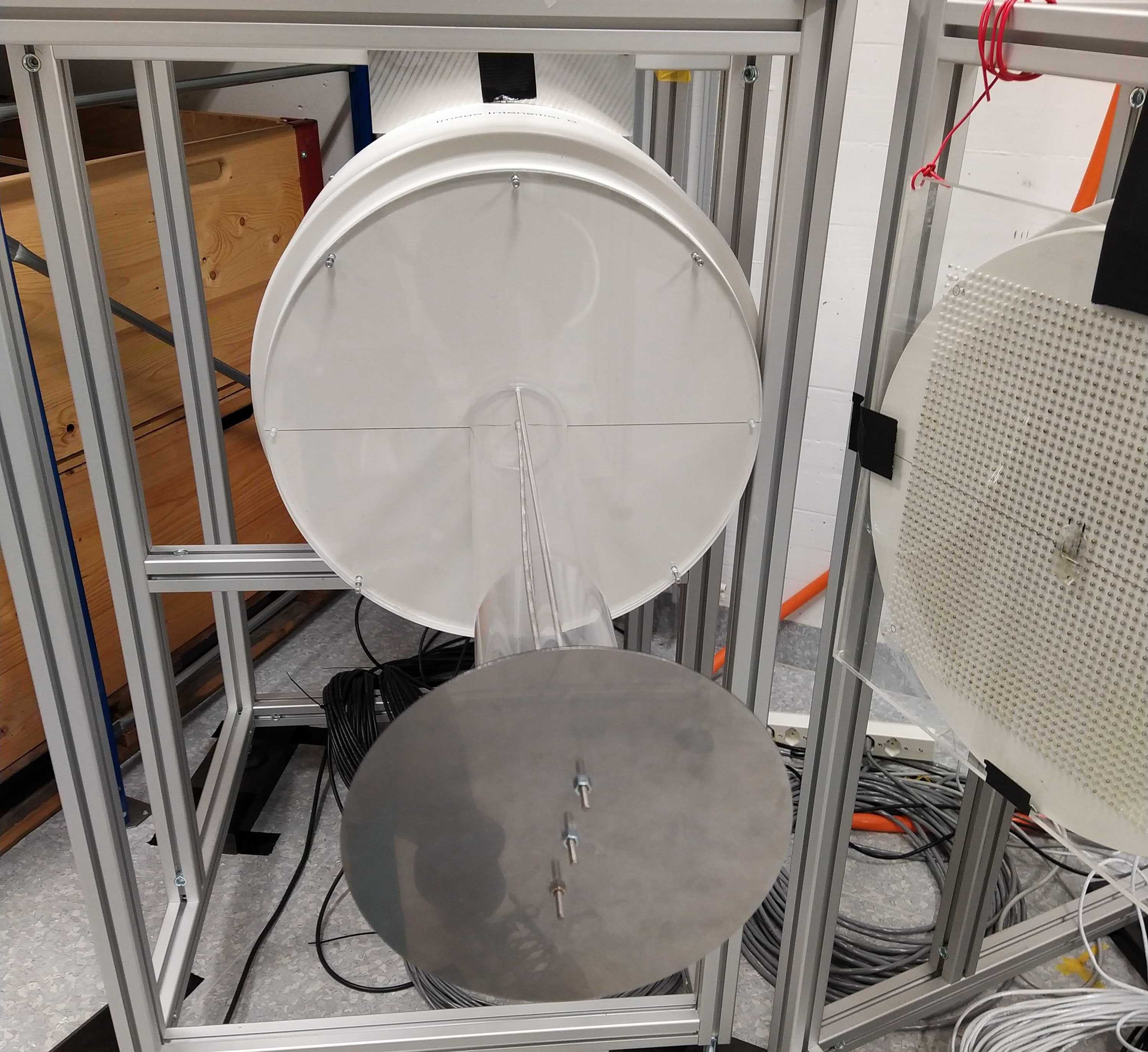}
        \caption{Tube calibration object.}
        \label{fig:tube}
    \end{subfigure}
    \caption{Calibration objects used in the dual-plane X-ray setup.}
    \label{fig:calibration_combined}
\end{figure}

\subsection{Dual-plane fluoroscope and X-ray system}

The imaging system comprised two X-ray tubes, two collimators, and two X-ray image intensifiers connected to high-speed cameras, all constructed in a bi-planar configuration. The two X-ray planes were mounted at an angle of 110° ± 0.5°, with source-intensifier distances of 1850 and 1855 mm, and intensifier sizes of 360 mm. Image acquisition was performed at a tube voltage of 40 kV and a tube current of 25 mA on a large focus. Each fluoroscopic image had a resolution of 1664 × 1600 pixels. 


\subsection{Data acquisition and implant design}
\begin{figure}[htb]
    \centering

    \begin{subfigure}{0.24\linewidth}
        \centering
        \includegraphics[width=\linewidth]{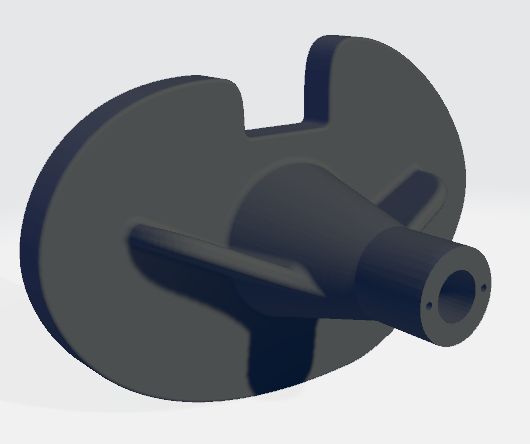}
        \caption{}
        \label{fig:tibia_asym_round_fixed}
    \end{subfigure}
    \hfill
    \begin{subfigure}{0.24\linewidth}
        \centering
        \includegraphics[width=\linewidth]{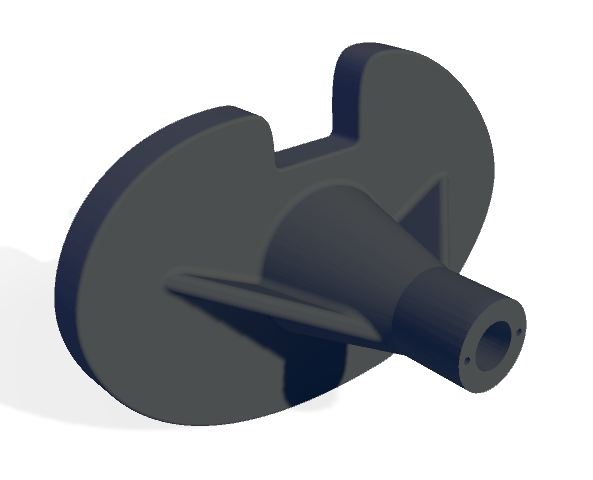}
        \caption{}
        \label{fig:tibia_asym_straight_fixed}
    \end{subfigure}
    \hfill
    \begin{subfigure}{0.24\linewidth}
        \centering
        \includegraphics[width=\linewidth]{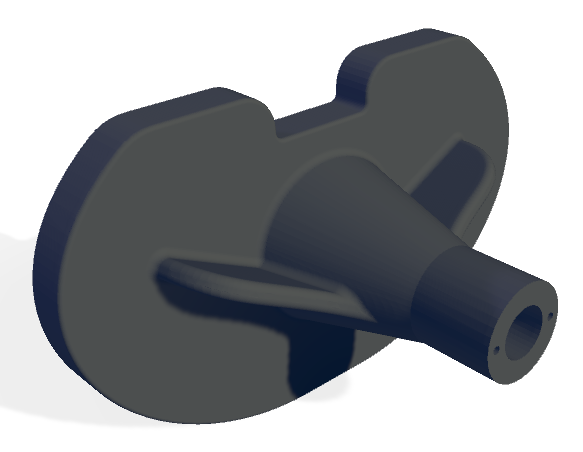}
        \caption{}
        \label{fig:tibia_sym_round_fixed}
    \end{subfigure}
    \hfill
    \begin{subfigure}{0.24\linewidth}
        \centering
        \includegraphics[width=\linewidth]{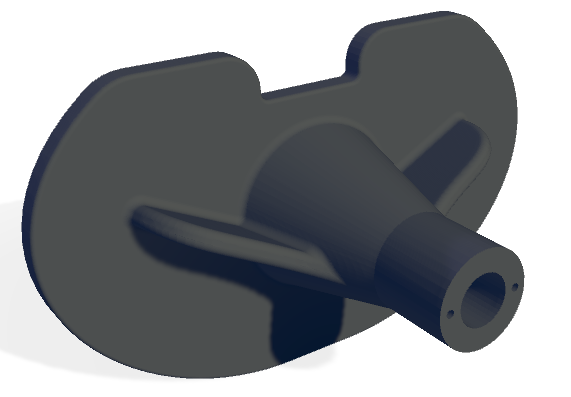}
       \caption{}
        \label{fig:tibia_sym_round_mobile}
    \end{subfigure}

    \begin{subfigure}{0.32\linewidth}
        \centering
        \includegraphics[width=\linewidth]{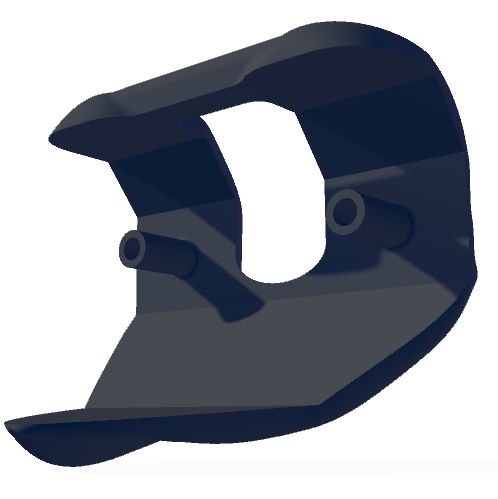}
        \caption{}
        \label{fig:femur_crossbar}
    \end{subfigure}
    \hfill
    \begin{subfigure}{0.32\linewidth}
        \centering
        \includegraphics[width=\linewidth]{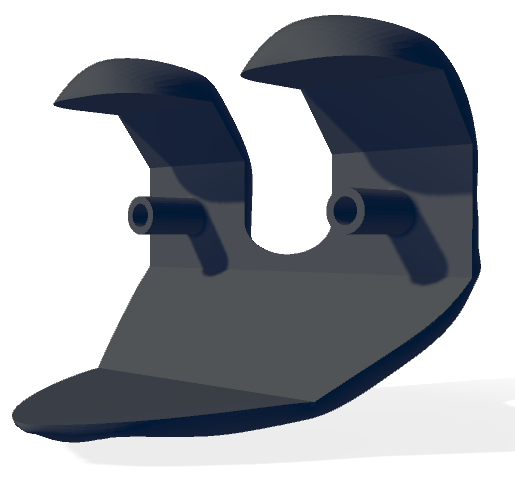}
        \caption{}
        \label{fig:femur_normal}
    \end{subfigure}
    \hfill
    \begin{subfigure}{0.32\linewidth}
        \centering
        \includegraphics[width=\linewidth]{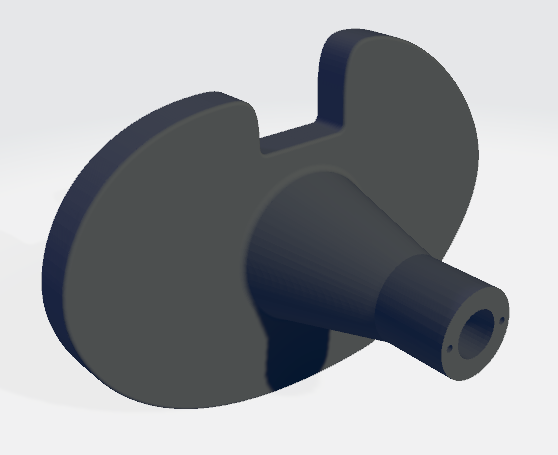}
        \caption{}
        \label{fig:tibia_asym_no_wings_fixed}
    \end{subfigure}

    \caption{Comparison of different tibial and femoral implant designs: 
    (a) Asymmetric round wings with fixed bearing, 
    (b) Asymmetric straight wings with fixed bearing, 
    (c) Symmetric round wings with fixed bearing,
    (d) Symmetric round wings with mobile bearing,
    (e) Femur crossbar design,
    (f) Femur normal design,
    (g) Asymmetric no-wings with fixed bearing.}
    \label{fig:implant_designs}
\end{figure}

The study included two femoral and five tibial radiopaque implant designs manufactured from aluminium to mimic widely used geometries (Fig. \ref{fig:implant_designs}). The implant geometries were adapted from CAD models provided by Zimmer Biomet (Winterthur, Switzerland) by smoothing all intricate geometries and adjusting non-load support wings while retaining the overall topology. A total of ten femur-tibia implant combinations were tested, with each combination following kinematic trajectories derived from the CAMS-Knee datasets \cite{Taylor2017,Dreyer2022}. The implant poses for each time instant replicated the complete trials of various activities. 

Each component was securely mounted to the end-effector of an industrial-grade six-axis precision robot (HORST600, Fruitcore Robotics GmbH, Konstanz, Germany), operating with a reported precision of ±0.05 mm. An adapter ensured proper positioning and prevented interference between the implant and the X-ray intensifier. The robotic system sequentially moved the implant to the predefined poses, holding each position for 500 ms before acquiring an X-ray image. This delay ensured that the end-effector remained stationary during imaging.

Data acquisition spanned ten measurement days, with one implant combination measured per day. Femur and tibia components were imaged separately to enable independent analysis as well as together to simulate complete knee implant configurations.

\subsection{Calibration measurements}
Prior to measurements, calibration procedures were performed to correct imaging distortions and establish accurate geometric relationships between the X-ray sources and intensifiers. Two key calibration processes were performed: distortion calibration (DISCAL) and source-intensifier calibration (SICAL).

For distortion correction, a bead grid phantom (45 × 45 beads with 7 mm spacing) was affixed to the intensifier surface (Fig. \ref{fig:bead_grid}). The known geometry of the bead grid enabled precise distortion correction using a coherent point drift method to establish correspondences, followed by third-order polynomial fitting with Powell optimisation to compute the transformation. 

For source-intensifier calibration (SICAL), a tube phantom was positioned 500 mm in front of the intensifier. This phantom featured a 200 mm diameter metal plate, concentrically aligned with the intensifier (Fig.\ref{fig:tube}), allowing accurate geometric calibration of the imaging system. The relative position of the X-ray source was determined by analyzing the projected shadows of the circular metal plate. 

\begin{figure}[htb]
    \centering

    \begin{subfigure}{0.4\linewidth}
        \centering
        \includegraphics[width=\linewidth]{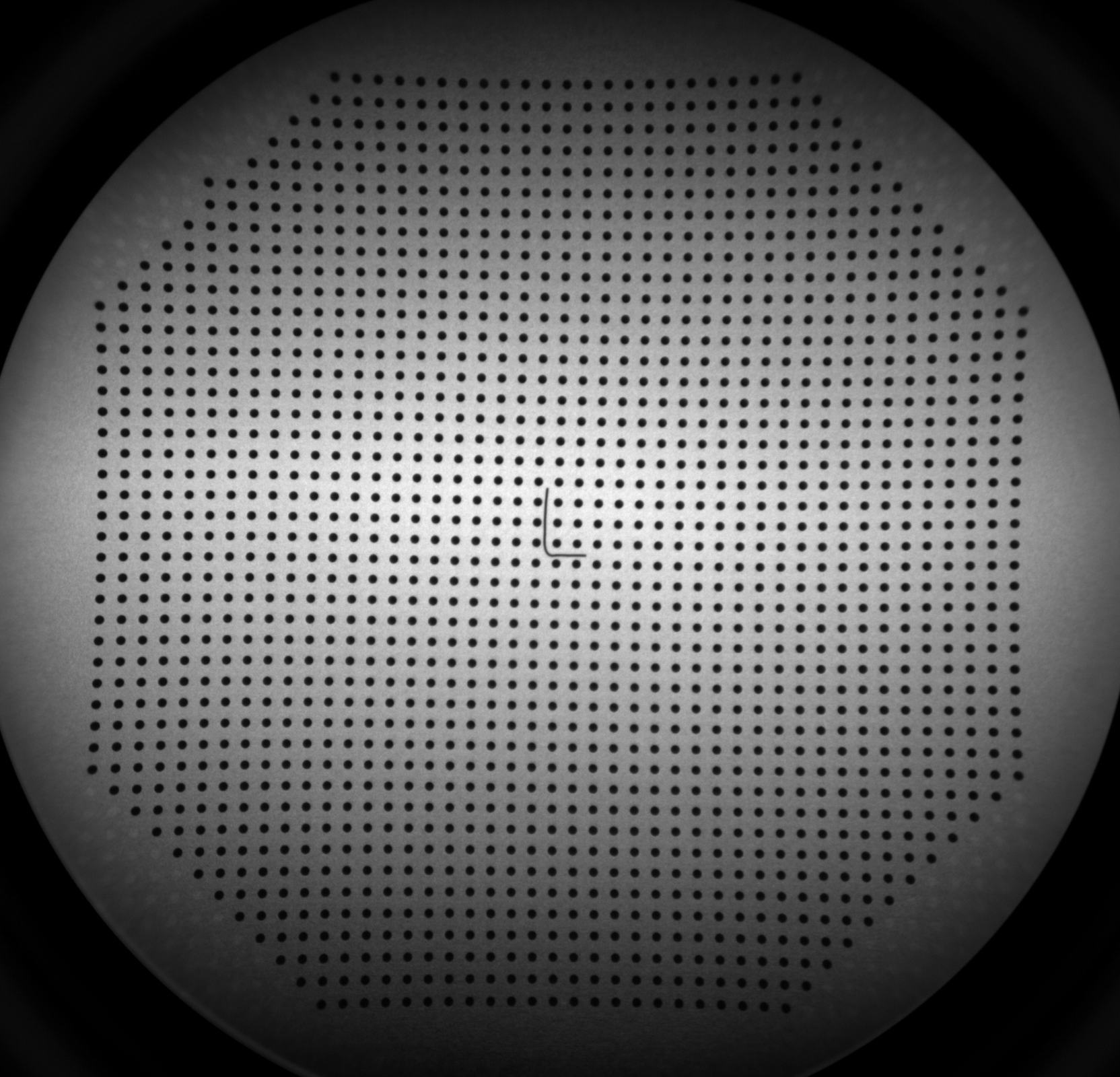}
        \caption{DISCAL grid - Camera A}
        \label{fig:discalA}
    \end{subfigure}
    \quad
    \begin{subfigure}{0.4\linewidth}
        \centering
        \includegraphics[width=\linewidth]{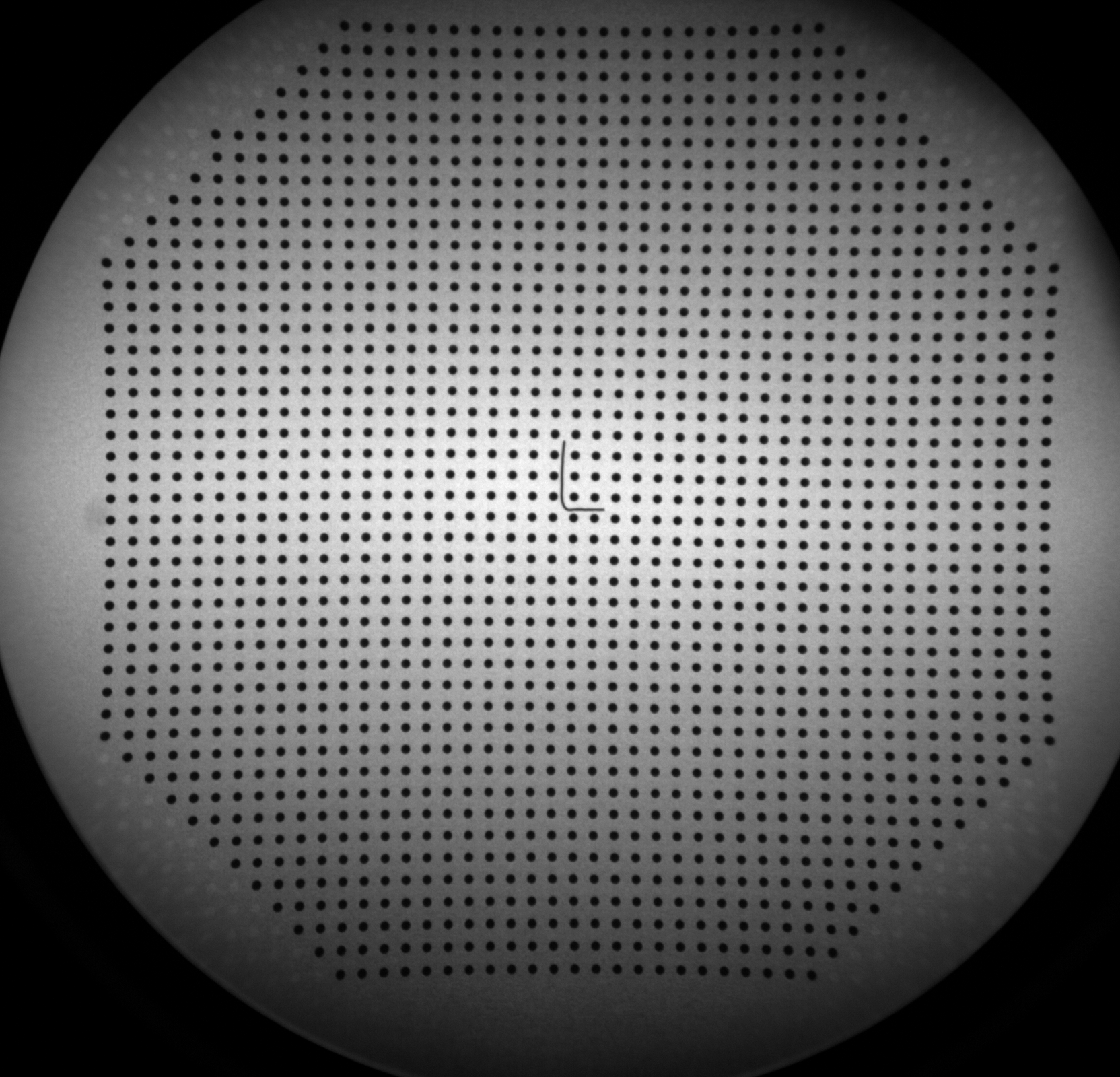}
        \caption{DISCAL grid - Camera B}
        \label{fig:discalB}
    \end{subfigure}

    \begin{subfigure}{0.4\linewidth}
        \centering
        \includegraphics[width=\linewidth]{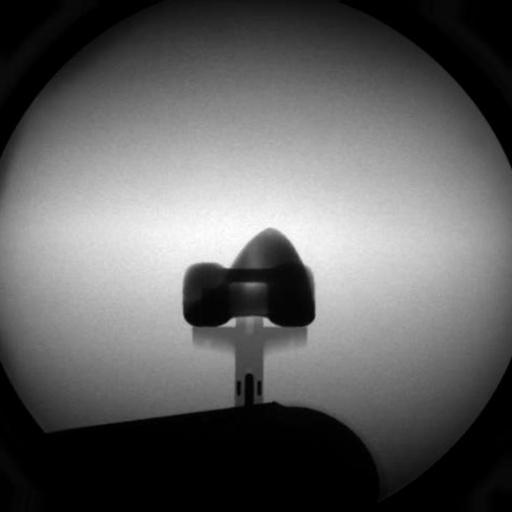}
        \caption{Default pose - Camera A}
        \label{fig:calibrated_bs_000001}
    \end{subfigure}
    \quad
    \begin{subfigure}{0.4\linewidth}
        \centering
        \includegraphics[width=\linewidth]{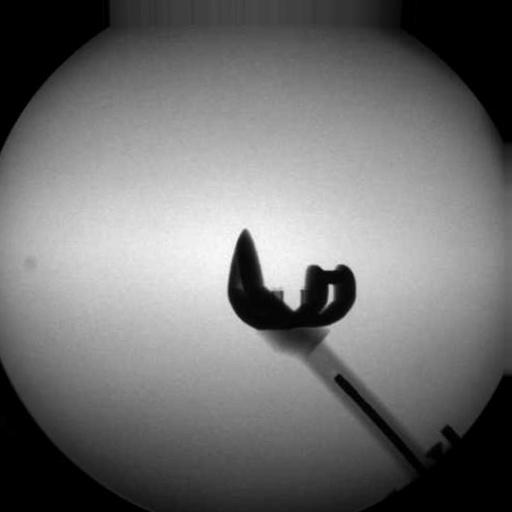}
        \caption{Default pose - Camera B}
        \label{fig:calibrated_fs_000001}
    \end{subfigure}

    \caption{Calibration images used in the experimental setup.}
    \label{fig:calibration_images}
\end{figure}

All resulting DISCAL and SICAL values, as well as the raw distortion grid images (Fig.\ref{fig:discalA}\&\ref{fig:discalB}), phantom measurements, and precomputed correction functions, were stored and are available within the Veriserum database. The database also includes code for computing and applying DISCAL and SICAL functions. While precomputed calibration values are available in the database for immediate application, users are welcome to reprocess the calibration parameters if necessary.

\subsection{Poses}
A total of 110,990 implant poses were recorded, covering movements from level walking (28k), stair descent (25k), ramp descent (14k), and chair sitting (42k) activities. Poses were derived from the CAMS-Knee kinematics database. Each pose corresponded to a predefined end-effector position provided to the robotic system.

Despite high robotic precision, minor deviations in the accuracy of reproducing the target pose occurred due to inherent robotic control system limitations. An automated image registration pipeline (Section 3.2) estimated the true implant pose by finding the rigid transformation between the rendered implant projection and the acquired fluoroscopic image. Additionally, a subset of 200 images was manually registered using a self-designed image registration software, ensuring correct calibrations. These manually registered poses served as a baseline for evaluating the automated registration performance. 


\section{Experiments and Results}
The published dataset includes X-ray images with corresponding target robot poses (110k), corresponding automatically registered poses for all images, and manually registered poses for a subset of 200 randomly selected images.

\subsection{Data Collection and Repeatability}
Data was collected over ten days, each day dedicated to a different femur-tibia implant combination. As part of quality assurance to check precision, the robotic system repositioned each implant to its default pose between each trial (Fig.\ref{fig:calibrated_bs_000001}\&\ref{fig:calibrated_fs_000001}). The average 2D correlations between the default pose images were computed using Normalized Cross Correlation (NCC) to ensure the repeatability of the data ~\cite{Grupp2018}. Across all measurement days, the correlation scores consistently exceeded 0.985, demonstrating a high inter-day positional repeatability of the robotic system. However, despite this high precision, discrepancies remained between the programmed target robot pose and the true implant pose in the fluoroscopic images due to minor inaccuracies inherent in the robotic control system (Fig. \ref{fig:error_percentiles}).

\subsection{Automated Pose Registration}

We developed a novel differentiable surface renderer using PyTorch3D for automated implant registration~\cite{Liu2019,Ravi2020}. This module is compatible with PyTorch, supports gradient-based optimization, and is designed for future integration with AI-based networks. The open-source code is publicly available within the Veriserum dataset to support the research community.

Since the rendered projections of the initial robot pose did not fully align with the fluoroscopic images, estimation of the true implant pose was necessary for accurate dataset annotation. Our differentiable renderer was used together with our automated registration method to refine the initial robot-based positioning and obtain the matched pose.

Each robot pose was optimised using the ADAM optimiser with NCC as the similarity metric, over 200 steps and a learning rate of 0.25. To assess registration accuracy, meticulously manually registered poses served as the gold standard reference. Translation was evaluated using in-plane L1 loss function, and rotation via the geodesic angle that measures the shortest path between two transformations on the rotation manifold.

\begin{figure}[htb]
    \centering
    \includegraphics[width=0.95\linewidth]{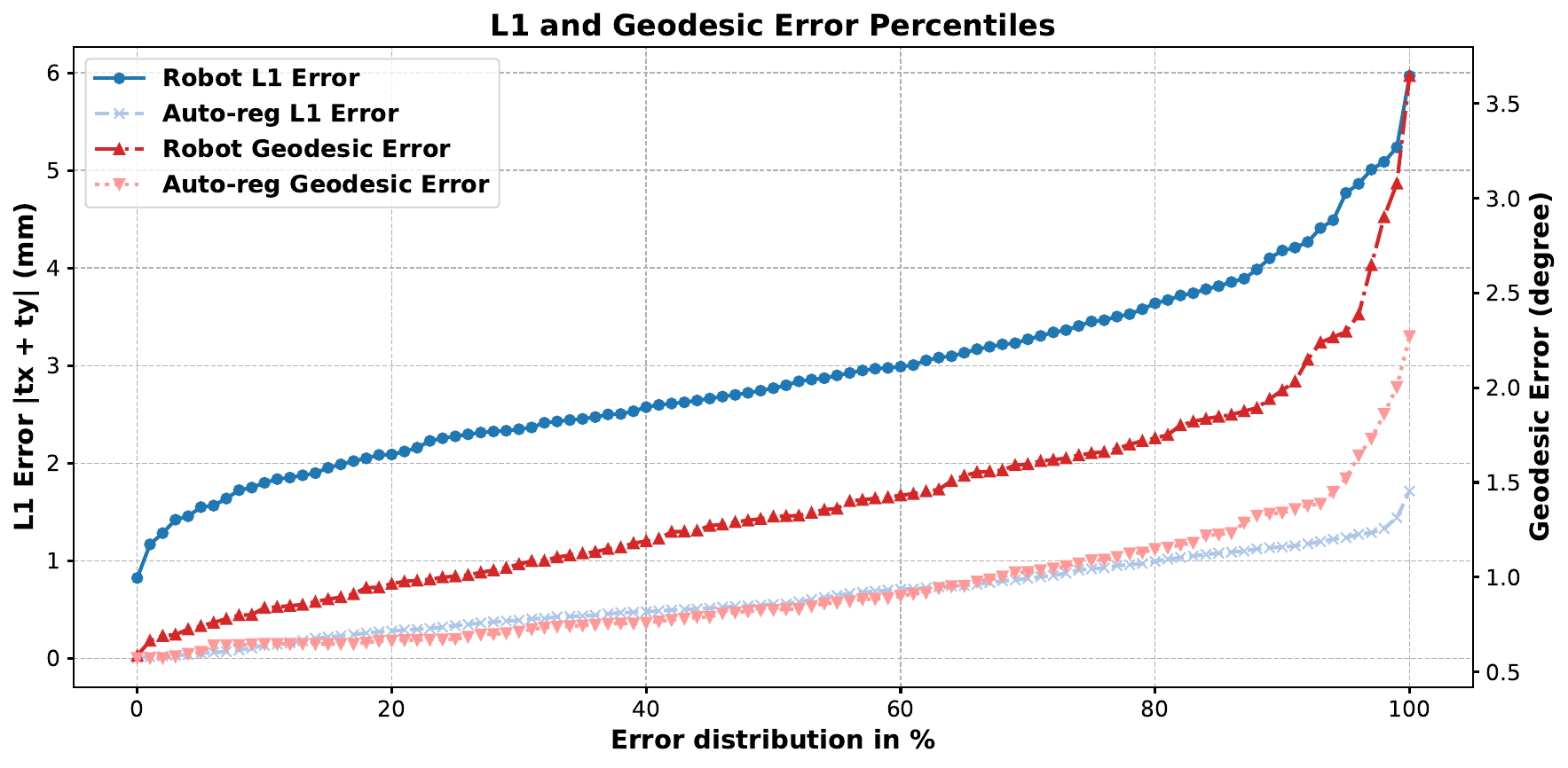}
    \caption{Error percentile plot of target robot poses and automatically registered poses (Auto-reg), compared to manual ground truth.}
    \label{fig:error_percentiles}
\end{figure}

 In approximately 65\% of the datasets, the error difference between the target robot pose and the manually registered ground truth exceeded 2.5 mm in L1 translational error and 1° in geodesic rotational error (Fig.\ref{fig:error_percentiles}). In contrast, the automatically registered poses achieved submillimeter accuracy compared to ground truth, with over 80\% images showing errors below 1 mm and 1°. This result demonstrates that the automated registration pipeline significantly improves pose accuracy (MAE < 0.8 mm, 0.9°), producing results comparable to time-intensive manual registration while offering a scalable and reproducible solution. Once this dataset is openly available to the community, both manually registered poses and automated registered poses can serve as ground truth for deep learning training. 

\section{Conclusion and Discussion}
This study presents Veriserum, the first open-source, high-quality dual-plane fluoroscopic dataset for knee implant analysis. It includes fluoroscopic images of diverse, clinically relevant poses, with both automated and manually registered pose annotations obtained through systematic calibration, making it a strong benchmark for evaluating 2D/3D image registration. Its pose diversity promotes robustness for enhancing data-driven methods~\cite{Wang2021,Li2018,Vogl2022}, and segmentation masks generated via rendering can assist implant segmentation and pose initialization~\cite{Jensen2023}. Veriserum, therefore, supports downstream tasks such as image registration, implant segmentation, and 3D reconstruction in post-operative patients~\cite{mescheder2019occupancynetworkslearning3d}. 

While Veriserum provides a valuable resource for medical imaging research, the absence of soft-tissue structures and real patient anatomy may limit the generalisability of deep learning models trained using these datasets. Additionally, since Veriserum time-series were reconstructed from datasets originally captured at only 25 or 30 Hz, its use in supporting time-series analysis maybe limited in some cases. Finally, the original fluoroscopic images were captured using analogue source-intensifier technology rather than digital flat panels, and hence spatial pixel errors were relatively large. Future work could expand the dataset to include natural bones with soft tissues, upgrade the X-ray system with a flat panel technology to avoid DISCAL-related errors, and increase the frequency of dynamic fluoroscopic sequences.

With its open availability, Veriserum is intended to advance biomechanics and AI-assisted fluoroscopic image analysis. As a comprehensive and reproducible benchmark, it enables development of automated 2D/3D registration, segmentation, and 3D reconstruction methods. As deep learning evolves, Veriserum represents a step towards bridging traditional model-based registration and AI-driven solutions.

    

\begin{credits}
\subsubsection{\ackname} 
The authors acknowledge the developers of the open-source tools used in this study, including PyTorch3D. Additionally, we would like to thank our collaborator HorstCosmos for providing the Fruitcore Robotics system used in the experiments.

\subsubsection{\discintname}
The authors declare no competing interests.

\end{credits}

%
%
%
\bibliographystyle{splncs04}
\bibliography{Paper-2716}

\begin{thebibliography}{10}
\providecommand{\url}[1]{\texttt{#1}}
\providecommand{\urlprefix}{URL }
\providecommand{\doi}[1]{https://doi.org/#1}

\bibitem{Chen2024}
Chen, M., Li, T., Zhang, Z., Kong, Y.: An optimization-based baseline for rigid 2d/3d registration applied to spine surgical navigation using cma-es  (2 2024), \url{http://arxiv.org/abs/2402.05642}

\bibitem{Dreyer2022}
Dreyer, M.J., Trepczynski, A., Nasab, S.H.H., Kutzner, I., Schütz, P., Weisse, B., Dymke, J., Postolka, B., Moewis, P., Bergmann, G., Duda, G.N., Taylor, W.R., Damm, P., Smith, C.R.: European society of biomechanics s.m. perren award 2022: Standardized tibio-femoral implant loads and kinematics. Journal of Biomechanics  \textbf{141} (8 2022). \doi{10.1016/j.jbiomech.2022.111171}

\bibitem{Flood2018}
Flood, P.D., Banks, S.A.: Automated registration of 3-d knee implant models to fluoroscopic images using lipschitzian optimization. IEEE Transactions on Medical Imaging  \textbf{37},  326--335 (1 2018). \doi{10.1109/TMI.2017.2773398}

\bibitem{Gopalakrishnan2023}
Gopalakrishnan, V., Dey, N., Golland, P.: Intraoperative 2d/3d image registration via differentiable x-ray rendering  (12 2023), \url{http://arxiv.org/abs/2312.06358}

\bibitem{Grupp2018}
Grupp, R.B., Armand, M., Taylor, R.H.: Patch-based image similarity for intraoperative 2d/3d pelvis registration during periacetabular osteotomy. In: Lecture Notes in Computer Science (including subseries Lecture Notes in Artificial Intelligence and Lecture Notes in Bioinformatics). vol. 11041 LNCS, pp. 153--163. Springer Verlag (2018). \doi{10.1007/978-3-030-01201-4_17}

\bibitem{Grupp2020}
Grupp, R.B., Unberath, M., Gao, C., Hegeman, R.A., Murphy, R.J., Alexander, C.P., Otake, Y., McArthur, B.A., Armand, M., Taylor, R.H.: Automatic annotation of hip anatomy in fluoroscopy for robust and efficient 2d/3d registration. International Journal of Computer Assisted Radiology and Surgery  \textbf{15},  759--769 (5 2020). \doi{10.1007/s11548-020-02162-7}

\bibitem{Guan2016}
Guan, S., Gray, H.A., Keynejad, F., Pandy, M.G.: Mobile biplane x-ray imaging system for measuring 3d dynamic joint motion during overground gait. IEEE Transactions on Medical Imaging  \textbf{35},  326--336 (1 2016). \doi{10.1109/TMI.2015.2473168}

\bibitem{Jensen2023}
Jensen, A.J., Flood, P.D., Palm-Vlasak, L.S., Burton, W.S., Chevalier, A., Rullkoetter, P.J., Banks, S.A.: Joint track machine learning: An autonomous method of measuring total knee arthroplasty kinematics from single-plane x-ray images. Journal of Arthroplasty  (10 2023). \doi{10.1016/j.arth.2023.05.029}

\bibitem{Li2018}
Li, Y., Wang, G., Ji, X., Xiang, Y., Fox, D.: Deepim: Deep iterative matching for 6d pose estimation  (3 2018). \doi{10.1007/s11263-019-01250-9}, \url{http://arxiv.org/abs/1804.00175 http://dx.doi.org/10.1007/s11263-019-01250-9}

\bibitem{Liu2019}
Liu, S., Li, T., Chen, W., Li, H.: Soft rasterizer: A differentiable renderer for image-based 3d reasoning  (4 2019), \url{http://arxiv.org/abs/1904.01786}

\bibitem{mescheder2019occupancynetworkslearning3d}
Mescheder, L., Oechsle, M., Niemeyer, M., Nowozin, S., Geiger, A.: Occupancy networks: Learning 3d reconstruction in function space (2019), \url{https://arxiv.org/abs/1812.03828}

\bibitem{Petersen2023}
Petersen, E.T., Vind, T.D., Jürgens-Lahnstein, J.H., Christensen, R., de~Raedt, S., Brüel, A., Rytter, S., Andersen, M.S., Stilling, M.: Evaluation of automated radiostereometric image registration in total knee arthroplasty utilizing a synthetic-based and a ct-based volumetric model. Journal of Orthopaedic Research  \textbf{41},  436--446 (2 2023). \doi{10.1002/jor.25359}

\bibitem{Postolka2020}
Postolka, B., Schütz, P., Fucentese, S.F., Freeman, M.A., Pinskerova, V., List, R., Taylor, W.R.: Tibio-femoral kinematics of the healthy knee joint throughout complete cycles of gait activities. Journal of Biomechanics  \textbf{110} (9 2020). \doi{10.1016/j.jbiomech.2020.109915}

\bibitem{Postolka2022}
Postolka, B., Taylor, W.R., List, R., Fucentese, S.F., Koch, P.P., Schütz, P.: Isb clinical biomechanics award winner 2021: Tibio-femoral kinematics of natural versus replaced knees – a comparison using dynamic videofluoroscopy. Clinical Biomechanics  \textbf{96} (6 2022). \doi{10.1016/j.clinbiomech.2022.105667}

\bibitem{Ravi2020}
Ravi, N., Reizenstein, J., Novotny, D., Gordon, T., Lo, W.Y., Johnson, J., Gkioxari, G.: Accelerating 3d deep learning with pytorch3d  (7 2020), \url{http://arxiv.org/abs/2007.08501}

\bibitem{}
Stentz-Olesen, K., Nielsen, E.T., de~Raedt, S., Jørgensen, P.B., Sørensen, O.G., Kaptein, B., Søballe, K., Stilling, M.: Reconstructing the anterolateral ligament does not decrease rotational knee laxity in acl-reconstructed knees. Knee Surgery, Sports Traumatology, Arthroscopy  \textbf{25},  1125--1131 (4 2017). \doi{10.1007/s00167-017-4500-3}

\bibitem{Stevsic2021}
Stevsic, S., Hilliges, O.: Spatial attention improves iterative 6d object pose estimation  (1 2021), \url{http://arxiv.org/abs/2101.01659}

\bibitem{Taylor2017}
Taylor, W.R., Schütz, P., Bergmann, G., List, R., Postolka, B., Hitz, M., Dymke, J., Damm, P., Duda, G., Gerber, H., Schwachmeyer, V., Nasab, S.H.H., Trepczynski, A., Kutzner, I.: A comprehensive assessment of the musculoskeletal system: The cams-knee data set. Journal of Biomechanics  \textbf{65},  32--39 (12 2017). \doi{10.1016/j.jbiomech.2017.09.022}

\bibitem{Tsai2011}
Tsai, T.Y., Lu, T.W., Kuo, M.Y., Lin, C.C.: Effects of soft tissue artifacts on the calculated kinematics and kinetics of the knee during stair-ascent. Journal of Biomechanics  \textbf{44},  1182--1188 (4 2011). \doi{10.1016/j.jbiomech.2011.01.009}

\bibitem{Vogl2022}
Vogl, F., Schutz, P., Postolka, B., List, R., Taylor, W.: Personalised pose estimation from singleplane moving fluoroscope images using deep convolutional neural networks. PLoS ONE  \textbf{17} (6 2022). \doi{10.1371/journal.pone.0270596}

\bibitem{Wang2021}
Wang, C., Xie, S., Li, K., Wang, C., Liu, X., Zhao, L., Tsai, T.Y.: Multi-view point-based registration for native knee kinematics measurement with feature transfer learning. Engineering  \textbf{7},  881--888 (6 2021). \doi{10.1016/j.eng.2020.03.016}

\bibitem{Zhang2023}
Zhang, B., Faghihroohi, S., Azampour, M.F., Liu, S., Ghotbi, R., Schunkert, H., Navab, N.: A patient-specific self-supervised model for automatic x-ray/ct registration. In: Lecture Notes in Computer Science (including subseries Lecture Notes in Artificial Intelligence and Lecture Notes in Bioinformatics). vol. 14228 LNCS, pp. 515--524. Springer Science and Business Media Deutschland GmbH (2023). \doi{10.1007/978-3-031-43996-4_49}

\end{thebibliography}
%




\end{document}